\documentclass[letter, 10 pt, conference]{ieeeconf}  

\IEEEoverridecommandlockouts 
\usepackage[english]{babel}
\usepackage[utf8]{inputenc}
\usepackage{times} 
\usepackage{cite}
\usepackage{balance}
\usepackage{footmisc}

\usepackage{hyperref}
\hypersetup{
    colorlinks=true,
    linkcolor=black,
    urlcolor=blue
}

\usepackage[nolist]{acronym}

\usepackage{multirow}

\usepackage{siunitx}
\usepackage{graphicx} 
\usepackage{epsfig} 
\usepackage{pgfplots}
\usepackage[font=small]{caption}
\usepackage[font=footnotesize]{subcaption}
\usepackage{pgfplots}

\usepackage{float}

\usepackage[export]{adjustbox}

\usepackage{booktabs}

\usepackage[markup=bfit, deletedmarkup=sout, authormarkup=brackets]{changes}
\definechangesauthor[name={Pavel S.}, color=blue]{PS}

\usepackage{mathtools}
\usepackage{nicefrac}
\usepackage{amsmath}
\usepackage{amssymb}
\usepackage{bm} 

\usepackage{algorithm}
\usepackage[noend]{algpseudocode}

\makeatletter
\def\BState{\State\hskip-\ALG@thistlm}
\makeatother

\usepackage{tikz}
\pgfdeclarelayer{foreground}
\pgfsetlayers{background, main, foreground}
\usetikzlibrary{quotes, angles, backgrounds, arrows, automata, shapes, positioning, calc, through, spy, decorations.pathreplacing, decorations.markings, arrows.meta, automata, petri}

\tikzset{
    imglabel/.style={
      rectangle,
      inner sep=2pt,
      text=black,
      minimum height=1em,
      text centered,
      fill=white,
      fill opacity=1.0,
      text opacity=1,
      anchor=south west,
    },
  }
\tikzset{
	state/.style={
		rectangle,
		draw=black, very thick,
		minimum height=1.0em,
		text centered,
	},
  smallstate/.style={
    rectangle,
    draw=black, very thick,
    minimum height=0.2em,
    text centered,
  },
}
\tikzset{
  on each segment/.style={
    decorate,
    decoration={
      show path construction,
      moveto code={},
      lineto code={
        \path [#1]
        (\tikzinputsegmentfirst) -- (\tikzinputsegmentlast);
      },
      curveto code={
        \path [#1] (\tikzinputsegmentfirst)
        .. controls
        (\tikzinputsegmentsupporta) and (\tikzinputsegmentsupportb)
        ..
        (\tikzinputsegmentlast);
      },
      closepath code={Acp
        \path [#1]
        (\tikzinputsegmentfirst) -- (\tikzinputsegmentlast);
      },
    },
  },
  mid arrow/.style={postaction={decorate,decoration={
        markings,
        mark=at position .5 with {\arrow[#1]{stealth}}
      }}},
}

\graphicspath{{./figures/}}

\let\oldtwocolumn\twocolumn
\renewcommand\twocolumn[1][]{%
    \oldtwocolumn[{#1}{
    \vspace{-9.5mm}
    \begin{center}
      \input{fig/TeX/uavs.tex}
      \vspace{-1mm}
      \captionof{figure}{MRS UAV Platforms used for diverse range of real-world applications (a)-(f) and their simulated variants (g)-(l).}
    \end{center}}
    ]
}

\newcommand\copyrighttext{%
	\small \begin{center} \color{red} \textcopyright\,2022 IEEE. Personal use of this material is permitted. Permission from IEEE must be obtained for all other uses, in any current or future media, including reprinting/republishing this material for advertising or promotional purposes, creating new collective works, for resale or redistribution to servers or lists, or reuse of any copyrighted component of this work in other works. \end{center}}
\newcommand\copyrightnotice{%
	\begin{tikzpicture}[remember picture,overlay]
	\node[anchor=south,yshift=25.6cm] at (current page.south) 
	{\color{red}\fbox{\parbox{\dimexpr\textwidth-\fboxsep-\fboxrule\relax}{\copyrighttext}}};
	\end{tikzpicture}%
}

\title{\copyrightnotice \LARGE \bf
    MRS Modular UAV Hardware Platforms for Supporting Research in Real-World Outdoor and Indoor Environments
}

\author{%
    Daniel Hert$^{1}$, Tomas Baca$^{1}$, Pavel Petracek$^{1}$, Vit Kratky$^{1}$, Vojtech Spurny$^{1}$, Matej Petrlik$^{1}$, Matous Vrba$^{1}$,
    \\David Zaitlik$^{1}$, Pavel Stoudek$^{1}$, Viktor Walter$^{1}$, Petr Stepan$^{1}$, Jiri Horyna$^{1}$, Vaclav Pritzl$^{1}$, Giuseppe Silano$^{1}$,%
    \\Daniel Bonilla Licea$^{1}$, Petr Stibinger$^{1}$, Robert Penicka$^{1}$, Tiago Nascimento$^{1}$, and Martin Saska$^{1}$
    \thanks{$^{1}$Authors are with the Faculty of Electrical Engineering, Czech Technical University in Prague, Czech Republic. The corresponding author email: {\tt\small martin.saska@fel.cvut.cz.}}
    \thanks{This work was partially funded by the CTU grant no. SGS20/174/OHK3/3T/13, by the Czech Science Foundation (GAČR) under research project no. 20-10280S, no. 20-29531S and no. 22-24425S, by TAČR project no. FW01010317, by the OP VVV funded project CZ.02.1.01/0.0/0.0/16 019/0000765 ``Research Center for Informatics", by the NAKI II project no. DG18P02OVV069, by the European Union's Horizon 2020 research and innovation programme AERIAL-CORE under grant agreement no. 871479, by the Defense Advanced Research Projects Agency (DARPA), and by the Technology Innovation Institute - Sole Proprietorship LLC, UAE. Furthermore, computational resources were supplied by the project ``e-Infrastruktura CZ" (e-INFRA LM2018140) provided within the program Projects of Large Research, Development and Innovations Infrastructures.}
}

\begin{document}

\maketitle
\thispagestyle{empty} 
\pagestyle{empty} 



%

  
\begin{abstract}

 This paper presents a family of autonomous \acp{UAV} platforms designed for a diverse range of indoor and outdoor applications.
 The proposed \ac{UAV} design is highly modular in terms of used actuators, sensor configurations, and even \ac{UAV} frames.
 This allows to achieve, with minimal effort, a proper experimental setup for single, as well as, multi-robot scenarios.
 Presented platforms are intended to facilitate the transition from simulations, and simplified laboratory experiments, into the deployment of aerial robots into uncertain and hard-to-model real-world conditions.
 We present mechanical designs, electric configurations, and dynamic models of the \acp{UAV}, followed by numerous recommendations and technical details required for building such a fully autonomous \ac{UAV} system for experimental verification of scientific achievements.
 To show strength and high variability of the proposed system, we present results of tens of completely different real-robot experiments in various environments using distinct actuator and sensory configurations.

\end{abstract}



\section{INTRODUCTION}

Multi-rotor \acp{UAV} are especially appealing for applications in cluttered workspaces, as they are able to perform complex maneuvers and fly close to obstacles in a relatively safe manner. 
Although research towards deployment of \ac{UAV} systems in these demanding conditions is enormous, most of the proposed approaches are verified by numerical simulations only, and do not respect requirements of real-world deployment.
In multi-UAV research, the proportion of systems that were designed and subsequently experimentally verified in real-world conditions is even lower.
Our recent survey of UAV control approaches~\cite{Tiago2019} shows that only 6\% of the methods had been verified experimentally in real-world conditions, 9\% in simplified laboratory experiments, 35\% in robotic simulators, and the rest of the 240 papers were concluded by numerical simulations only.

After years of research, the Multi-Robot Systems (MRS) group in Prague\footnote{\url{http://mrs.felk.cvut.cz}} designed a UAV platform based on field experience in various robotic applications (aerial swarms, aerial manipulation, motion planning, remote sensing, etc.). 
The MRS Hardware (HW) and Software (SW) UAV systems have provided support for state-of-the-art research that has resulted in dozens of publications by several research groups. 
All results have been supported by experimental verification in real conditions, which subsequently shaped the system into the current highly modular and unique form. 
To allow access of dozens research groups worldwide into the MRS system, the proposed platform is accompanied with an actively maintained and well-documented implementation on Github\footnote{\url{https://github.com/ctu-mrs/mrs_uav_system}}, including a realistic UAV simulation tool, various sensors, and localization systems. The system is from its beginning also intended to be an efficient educational tool and has enabled more than 300 bachelor, master, and PhD students from more than 100 research groups worldwide to conduct experiments in real outdoor and indoor conditions. Using the proposed concept, research groups can build hardware platforms equipped with the MRS UAV system on their own or with a support of UAV industry, simply by using the modular DroneBuilder web page\footnote{\url{https://dronebuilder.fly4future.com}}.

\subsection{Contributions beyond state-of-the-art}

Although numerous autonomous UAV platforms have been designed, new concepts of autonomous fixed wings UAVs \cite{Cao2018} and helicopter \cite{Stingu2009} platforms are still being proposed. The research of rotary-wing UAV platforms is even more abundant, as different hardware designs are required for newly appearing applications. 
Designing new UAV concepts still suffers from problems such as the long design period, high manufacture cost, and a difficulty of platform maintenance.
Those are the main obstacles for more frequent verification of scientific achievements in real conditions for which they are intended. 

A recent work of \cite{Guo2021} proposed a design method to obtain a lightweight and maintainable UAV frame using a configurable design. The work of Flynn \cite{Flynn2013} focused on the aspect of reducing cost for building and maintaining UAV platforms. Wang et al. \cite{Wang2021} focused on designing multi-robot systems using commercial benchmark platforms. Finally, the work of Schacht-Rodriguez et al. \cite{SchachtRodriguez2018} addressed the design, construction, and instrumentation of a UAV hexacopter experimental platform. 
Although all the above mentioned platforms try to propose a general-use hardware for research purposes, none of them has been thoroughly tested in real world environments.
Therefore, the aforementioned platforms cannot facilitate the desired minimization of the reality gap we are focusing on in this paper.

Although the software part of the MRS system has been partially described in numerous publications and summarised in \cite{baca2021mrs}, this is the first time the hardware part of the co-designed system is detailed, which may be even more valuable for researchers.
The contributions of this work for the robotic and UAV communities can be summarized as:
\begin{itemize}
    \item We propose a modular UAV platform that can be used in various applications with distinct actuator and sensor configurations, while minimizing the effort required for changing the UAV setup and for its maintenance. This is an important aspect mainly in initial stages of experimenting with new approaches due to a high probability of collisions.
    \item The MRS UAV platforms facilitate the transition from simulation, and simplified  laboratory  experiments, into  deployment  of aerial robots in real-world conditions with minimal sim-to-real gap.
    \item The system is intended to support initial steps of researchers and students from different scientific areas, in which the UAVs are necessary tool for experimental validation of proposed concepts, to enlarge the community of active users of fully autonomous UAVs.
\end{itemize}









\section{MECHANICAL DESIGN AND PROTOTYPING}




\subsection{Frames and propulsion system consideration}

The frame is the fundamental component of each UAV as it determines its final size and the maximum diameter of used propellers that are affecting overall payload, endurance, and enabled set of sensors and actuators. Generally, larger propellers spin slower and are more efficient than smaller propellers with faster spin. To achieve long flight times and higher payloads, larger propellers are preferred. However, smaller platforms are easier to transport and operate, and mainly their smaller size is more suited for environments with obstacles, for which small-scale multi-rotor UAVs are advantageous. When choosing the frame size a compromise has to be made between the final size of the platform and its propulsion efficiency.

The MRS platforms are based on three basic frame sets for most of the experimental verifications that are not supplemented by special designs for few applications that require unusual properties or physical interaction with environment. 

\subsection{Basic frames}

All three basic frames selected to support most of the research work being conducted recently have similar construction using four arms that are sandwiched between two central structural boards. The arms with attached motors are made from plastic or carbon fiber, and the structural boards from glass Reinforced Epoxy Laminate or carbon fiber.

The smallest of the selected frames is DJI F450, equipped with 2212 KV920 motors and plastic 9.4 inch propellers. The platform is powered by a 4S 6750\,mAh lithium polymer battery. This combination offers about 0.5\,kg of usable payload and flight times between 10-15 minutes, depending on the payload. This platform is mostly used for swarming research and experiments which do not require large payloads.

The medium selected frame is Holybro X500, equipped with 3510 KV700 motors, carbon fiber 13 inch propellers and either one or two 4S 6750\,mAh lithium polymer batteries. The propulsion system of our X500 is upgraded over the default kit provided by Holybro (2216 KV920 motors with 10 inch propellers), to increase the payload capacity and flight time. The X500 can carry 1.5\,kg of usable payload. If longer flight times are required, a second battery can be connected in parallel, enhancing the flight time to over 20 minutes even with full payload. This platform was used in the DARPA SubT Challenge, as it combines compact size, large payload and long flight time. In addition to the real DARPA SubT Challenge, this platform achieved the best performance also in the virtual DARPA SubT Challenge, where MRS X500 was used by all teams occupying the first six places in the final competition.

The largest selected frame is Tarot T650, equipped with 4114 KV320 motors, carbon fiber 15 inch propellers and a 6S 8000\,mAh lithium polymer battery. This is the heaviest standard MRS platform with payload capacity of 2.5\,kg. It was used in the MBZIRC 2020 competition, where large payloads were required for carrying heavy bricks~\cite{baca2020autonomous}, a ball catching net~\cite{vrba2020autonomous}, and water bags~\cite{walter2020fr}. 
This platform is also used in most of the projects that require UAV manipulation.

\subsection{Task-specific custom platforms}

For some applications, the standard frames are not suitable, as a combination of higher payloads and smaller size is required, or a non-square shape platform is preferable. Different shape may be required also in case of physical interaction with environment or due to a special end effector.

For example, for autonomous documentation of historical buildings~\cite{kratky2021documentation, petracek2020dronument}, a platform with high payload is required to carry a full-size camera on a gimbal, Ouster LiDAR for mapping, localization and collision avoidance, and propeller guards to protect the platform and its surroundings. Contrary, the platform has to be as small as possible to maintain its ability to fly in confined environments. In this case, a coaxial propulsion was used. Two motors with propellers are mounted above each other, which allows to mount twice as many motors without increasing the size of the platform. This increases the maximum thrust and therefore the maximum payload, but the coaxial propulsion setup is less efficient with ~20\% higher power~\cite{Bondyra2016PerformanceOC} to generate the same thrust as a standard propulsion setup. The platform is shown in Fig.~\ref{fig:dronument}.

A platform with similar requirements and construction was used for a fire-fighting application~\cite{spurny2020autonomous}, utilizing a compressed CO$_2$ based launcher, which is able to shoot a fire-fighting capsule (containing water and fire suppressant substances) through a window of a burning building. The long shape of the launcher became the basis of the entire frame, which was built around it. Coaxial propulsion was also used for this task, to minimize the size of the platform. This platform is shown in Fig.~\ref{fig:dofec}.

A prototype for a research of autonomous UAVs intended for onboard localization and elimination of non-authorised drones~\cite{stasinchuk2020multiuav} in protected no-fly zones is based on a Tarot T18 (see Fig.~\ref{fig:eagle}). This octacopter platform with 18 inch propellers is capable of lifting up to 10\,kg of payload, which is required to catch and carry other drones. The octacopter construction also brings redundancy, as the UAV can lose one of its motors and remain controllable, which is especially important in this application, where the probability of a collision with a target drone is high. 



\subsection{3D printed prototyping}

For development of custom parts and achieving the required modularity, the 3D printing technology is useful in the phase of initial prototyping, replication of verified UAVs, and mainly for platform maintenance even during experimental campaigns out of laboratory. This technology allows designing unique frame-sensor combinations and quick modifications of sensory attachment to adapt to the requirements of ongoing onsite research.
3D printed parts can hold UV lights, RGB cameras, LiDARs, as well as custom PCBs and on-board computers. In addition, some supporting mechanisms simplifying UAV operations for research purposes, such as a battery cage, are also made through 3D printing technology. During the design, stress simulation and shape optimization are used to achieve the best strength to weight ratio for the part.

Another important part of the UAV that is made by 3D printing is the set of custom legs. These are designed to be strong enough to support the UAV while stationary on the ground and simultaneously as light as possible. In addition, the legs were designed to act as a modular holder for sensors and other equipment. For example, a Basler camera and additional LED lights can be mounted on one or more legs to provide lighting~\cite{saska2017documentation, kratky2021exploration}.  Moreover, if a different size of legs is needed, the former design can be easily modified and extended even during experimental campaigns. An important part of the leg design that was motivated by experience in multiple experimental campaigns are carefully chosen weak spots.  Hence, in a case of an emergency landing the leg breaks in a specific spot, absorbing the impact energy and therefore protecting the more important (and expensive) core of the platform. Finally, each leg can be easily and quickly swapped for a new piece. For testing novel research concepts that cannot be in principle reliable at initial stages of development, such protection mechanisms are crucial.
This is mainly the case for multi-UAV experiments, where it is almost impossible to avoid UAV collisions.



\subsection{Sensory equipment}

When selecting usable sensors, the weight of the sensor plays one of the key roles. 
The lightest LiDARs, such as Garmin LIDAR-Lite used by most of the MRS platforms, measure only in one direction. The 2D RPLiDAR A3 spins a single laser beam to scan a plane in relatively small resolution and accuracy. The heaviest sensors used by MRS platforms are multi-beam Ouster and Velodyne 3D LiDARs that provide also the best performance in terms of precision and amount of data gathered. 
The choice of sensor  therefore depends on the carrying capacity of the drone and also the price that is significantly increasing from 1D to 3D solutions.

Lightweight sensors also include RGB and RGB-D cameras. 
Intel RealSense RGB-D cameras are one of the suitable choices for navigation and obstacle detection in outdoor environments.  RealSense cameras provide directly depth in image computed from a pair of infrared cameras. Their disadvantage is a strong GNSS signal interference and in the case of the D455 models also image degradation caused by sunlight, which restricts their usage to GNSS denied environment such as forest or indoor environments, where also the influence of sun is decreased. 

When choosing an RGB camera for autonomous flight, the global shutter option must be taken into account, as the drone frame vibrates and strong image distortion occurs when using a rolling shutter.
Another important aspect is the camera resolution and frame-rate. 
Although, the best camera resolution and frame-rate are preferred in general, processing large images in real-time with onboard computers is challenging and a delay in image processing pipeline can affect the control system.

In addition, sonars may be used for detecting the height of drones flying above the water surface, infrared and thermal cameras are suitable for detecting objects with temperature different from the background, RGB and UV sensitive cameras for detecting neighboring drones in UAV teams, TimePix radiation sensors for radioactive objects tracking, and many other sensors of specific phenomena can be used for autonomous flight in special conditions and applications. 

Sensors that are used onboard of the MRS UAV platforms are as follows:

\begin{itemize}
    \item Rangefinder: Garmin LiDAR lite V3, MB1340 MaxBotix ultrasound, 
    \item Planar LiDAR: RPLiDAR A3,
    \item 3D LiDAR: Ouster OS1 and OS0 series, Velodyne VLP16,
    \item Cameras: Basler Dart daA1600, Bluefox MLC200w (grayscale or RGB),
    \item RGB-D cameras: Intel RealSense D435i and D455,
    \item GNSS: NEO-M8N and RTK Emlid Reach M2,
    \item Thermal camera: FLIR Lepton,
    \item Pixhawk sensors: gyroscopes, barometers, accelerometers (also available as separate sensors with better performance),
    \item UltraViolet Direction And Ranging (UVDAR)~\cite{walter2019uvdar}.
\end{itemize}

A key property of the MRS system is multi-sensor fusion to achieve reliable recognition, navigation and localization. Each sensor has its advantages and disadvantages, and with data fusion it is possible to obtain results in orders of magnitude better than using separate sensor data.

\begin{figure}[tb]
    \centering
    \includegraphics[trim={0mm 40mm 0mm 110mm},clip,width=\columnwidth]{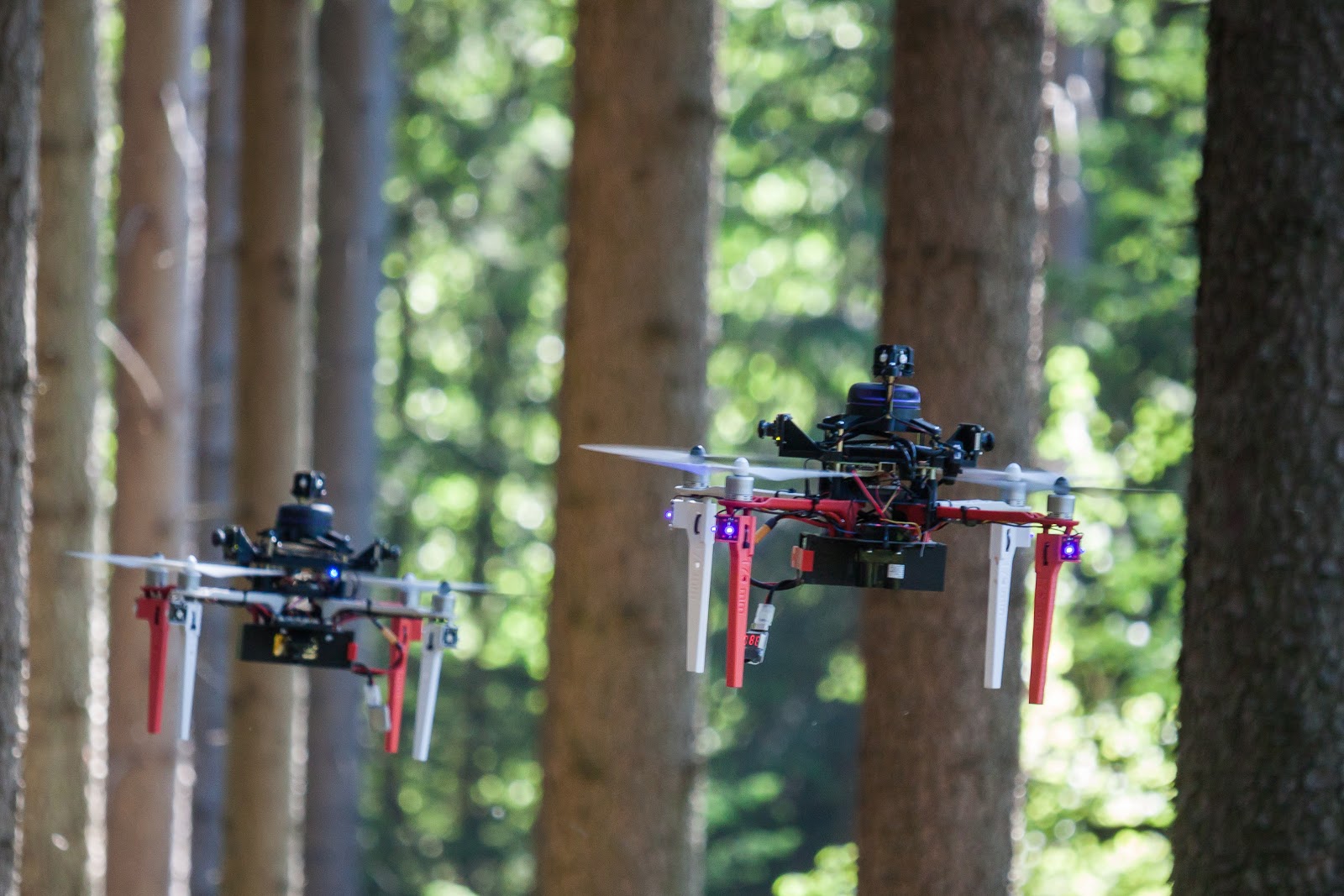}
    \vspace{-5mm}
    \caption{A pair of DJI F450-based UAV platforms equipped with the UVDAR system. Note the ultraviolet LEDs and dual cameras on the sides of the UAV body.}
    \label{fig:uvdar_hw}
    \vspace{-1em}
\end{figure}

\begin{figure}[tb]
    \centering
    \includegraphics[trim={0mm 40mm 60mm 40mm},clip,width=\columnwidth]{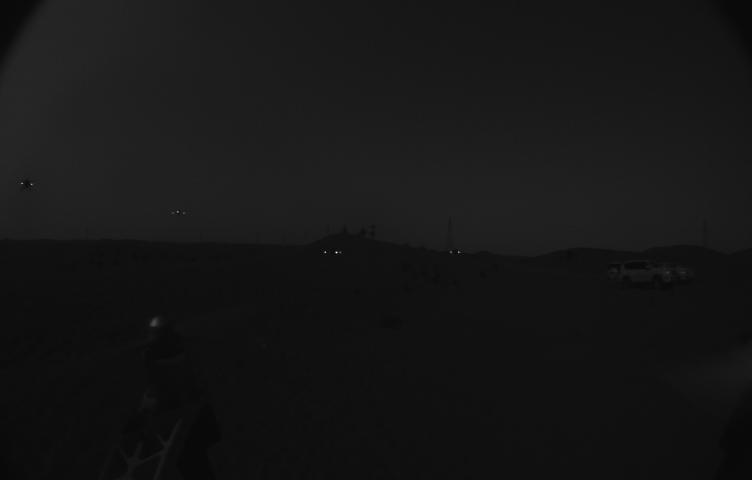}
    \vspace{-5mm}
    \caption{View from onboard ultraviolet-sensitive cameras used in the UVDAR system. Note the clarity of the onboard LED markers of neighboring UAVs, despite this image being captured at midday in a desert setting.}
    \label{fig:uvdar_hw2}
    \vspace{-1em}
\end{figure}

When deploying multiple UAVs, the MRS platforms have to be able to fly and cooperate even when direct radio communication is not reliable and external localization systems are not available.
Such cooperation requires that the agents in a swarm or formation can localize each other based on their onboard sensory equipment.
Computer vision is a state-of-the-art paradigm to mutual relative localization of robots~\cite{swarm_survey,vrba2019onboard,vrba2020markerless}. 
However, most of its current implementations suffer from degraded performance in general outdoor and indoor conditions, particularly when lighting conditions are concerned, in addition to significant computational complexity when payload of UAV is limited.
One solution used by the MRS UAV platforms is an optional smart sensor, called the UVDAR system~\cite{walter2019uvdar} shown in Figs~\ref{fig:uvdar_hw} and~\ref{fig:uvdar_hw2}.

This open-source\footnote{\url{https://github.com/ctu-mrs/uvdar_core}} system enables robust mutual localization based on active ultraviolet markers broadcasting identification signals, that are detected and decoded using ultraviolet-sensitive cameras ignoring significant portion of the image background \cite{walter2018mutual,walter2018fast}.
This method has been tested with great success in multiple real-world deployments of swarms and formations of UAVs \cite{petracek2020bioinspired,dmytruk2020safe,novak2021fast,afzal2021Icra,walter2019uvdar,midgard,Horyna2022ICUAS}, and it is being continuously developed and improved-upon. 



\begin{figure*}[!t]
    \centering
    \resizebox{1.0\textwidth}{!}{
    \pgfdeclarelayer{foreground}
\pgfsetlayers{background,main,foreground}

\makeatletter
\newcommand{\gettikzxy}[3]{%
  \tikz@scan@one@point\pgfutil@firstofone#1\relax
  \edef#2{\the\pgf@x}%
  \edef#3{\the\pgf@y}%
}
\makeatother

\tikzset{radiation/.style={{decorate,decoration={expanding waves,angle=90,segment length=4pt}}}}

\begin{tikzpicture}[->,>=stealth', node distance=3.0cm,scale=1.0, every node/.style={scale=1.0}]


  \node[state, shift = {(0.0, 0.0)}] (battery) {
      \begin{tabular}{c}
        \footnotesize Battery
      \end{tabular}
    };

  \node[state, right of = battery, shift = {(0.5, 0.0)}] (measurement) {
      \begin{tabular}{c}
        \footnotesize U and I\\
        \footnotesize measurement
      \end{tabular}
    };
    
  \node[state, right of = measurement, shift = {(0.5, 0.0)}] (main_psu) {
      \begin{tabular}{c}
        \footnotesize Main computer\\
        \footnotesize power supply
      \end{tabular}
    };
        
  \node[state, right of = main_psu, shift = {(0.5, 0.0)}] (FT4232H) {
      \begin{tabular}{c}
        \footnotesize FT4232H
      \end{tabular}
    };
            
  \node[state, right of = FT4232H, shift = {(0.5, 0.0)}] (5v) {
      \begin{tabular}{c}
        \footnotesize 3x 5V\\
        \footnotesize power supply
      \end{tabular}
    };
            
  \node[state, right of = 5v, shift = {(0.5, 0.0)}] (plant) {
      \begin{tabular}{c}
        \footnotesize UAV plant\\
        \footnotesize ESCs, motors
      \end{tabular}
    };
 
  \node[state, above of = main_psu, shift = {(0.0, -0.5)}] (main_computer) {
      \begin{tabular}{c}
        \footnotesize Main computer
      \end{tabular}
    };

  \node[state, right of = main_computer, shift = {(0.5, 0.0)}] (mrs_modules) {
      \begin{tabular}{c}
        \footnotesize 3x MRS modules 
      \end{tabular}
    };
    
  \node[state, right of = mrs_modules, shift = {(0.5, 0.0)}] (pixhawk) {
      \begin{tabular}{c}
        \footnotesize Pixhawk 
      \end{tabular}
    };


  
 \path[->, line width=0.5mm] ($(battery.east) + (0.0, 0)$) edge [] node[above, midway, shift = {(0.1, 0.05)}] {
   \begin{tabular}{c}
 \end{tabular}} ($(measurement.west) + (0.0, 0.00)$);

 \draw[-, line width=0.5mm] ($(measurement.south)+(0.0, 0.0)$) -- ($(measurement.south) + (0.0, -0.5)$) node[right, midway, shift = {(0.5, 0.0)}] {
    \begin{tabular}{c}
        \footnotesize Battery power 
 \end{tabular}} -- ($(plant.south) + (0.45, -0.5)$) edge [->] ($(plant.south)+ (0.45, -0.0)$);
 \path[->, line width=0.5mm] ($(plant.south) + (0.15, -0.5)$) edge ($(plant.south) + (0.15, 0.0)$);
 \path[->, line width=0.5mm] ($(plant.south) + (-0.15, -0.5)$) edge ($(plant.south) + (-0.15, 0.0)$);
 \path[->, line width=0.5mm] ($(plant.south) + (-0.45, -0.5)$) edge ($(plant.south) + (-0.45, 0.0)$);
    
  \path[->, line width=0.5mm] ($(main_psu.south) + (0.0, -0.5)$) edge ($(main_psu.south) + (0.0, 0.0)$);
  \path[->, line width=0.5mm] ($(5v.south) + (0.0, -0.5)$) edge ($(5v.south) + (0.0, 0.0)$);
  \path[->, line width=0.5mm] ($(main_psu.north) + (-0.4, 0.0)$) edge ($(main_computer.south) + (-0.4, 0.0)$);
  
 \draw[<-] ($(main_computer.south) + (0.4, 0.0)$) -- node[right, near start, shift = {(-0.2, -0.1)}] {
    \begin{tabular}{c}
        \footnotesize USB 
 \end{tabular}} ($(main_computer.south |- FT4232H.north) + (0.4, 0.6)$) -- ($(FT4232H.west |- FT4232H.north) + (-0.6, 0.6)$)  -- ($(FT4232H.west) + (-0.6, .0)$) edge [->] ($(FT4232H.west) + (0.0, 0.0)$);
 
   \path[<->] ($(FT4232H.north) + (-0.3, 0.0)$) edge node[left, near end, shift = {(0.2, 0.05)}] {
    \begin{tabular}{c}
        \footnotesize UART 
 \end{tabular}} ($(mrs_modules.south) + (-0.3, 0.0)$);
   \path[<->] ($(FT4232H.north) + (-0.1, 0.0)$) edge ($(mrs_modules.south) + (-0.1, 0.0)$);
   \path[<->] ($(FT4232H.north) + (0.1, 0.0)$) edge ($(mrs_modules.south) + (0.1, 0.0)$);
   
    \gettikzxy{(pixhawk.south)}{\psx}{\psy}
    \gettikzxy{(FT4232H.north)}{\fnx}{\fny}
   \draw[<-] ($(FT4232H.north) + (0.3, 0.0)$) -- ($(\fnx,\psy) + (0.3, -0.45)$) -- ($(pixhawk.south) + (-0.45, -0.45)$) edge [->] ($(pixhawk.south) + (-0.45, 0.0)$);
   
    \path[->, line width=0.5mm] ($(5v.north) + (0.15, 0.0)$) edge ($(pixhawk.south) + (0.15, 0.0)$);
    \path[->, line width=0.5mm] ($(5v.north) + (-0.15, 0.0)$) edge ($(pixhawk.south) + (-0.15, 0.0)$);
    \draw[-, line width=0.5mm] ($(5v.north) + (-0.45, 0.0)$) -- ($(5v.north |- mrs_modules.south) + (-0.45, -1.2)$) -- ($(mrs_modules.south) + (0.8, -1.2)$) edge [->] ($(mrs_modules.south) + (0.8, 0.0)$);
    
    \draw[-] ($(pixhawk.south) + (0.45, 0.0)$)  -- node[right, near start, shift = {(-0.2, -0.1)}] {
    \begin{tabular}{c}
        \footnotesize PWM/Dshot 
 \end{tabular}}($(pixhawk.south |- plant.east) + (0.45, 1.0)$) -- ($(pixhawk.south |- plant.east) + (1.45, 1.0)$) -- ($(pixhawk.south |- plant.west) + (1.45, -0.3)$) edge [->] ($(plant.west) + (0.0, -0.3)$);
 
  \path[->] ($(pixhawk.south |- plant.west) + (1.45, -0.1)$) edge [->] ($(plant.west) + (0.0, -0.1)$);
  \path[->] ($(pixhawk.south |- plant.west) + (1.45, 0.1)$) edge [->] ($(plant.west) + (0.0, 0.1)$);
  \path[->] ($(pixhawk.south |- plant.west) + (1.45, 0.3)$) edge [->] ($(plant.west) + (0.0, 0.3)$);
    


  \begin{pgfonlayer}{background}
    \path (measurement.west |- measurement.north)+(-0.45,0.8) node (a) {};
    \path (measurement.south -| 5v.east)+(0.75,-0.80) node (b) {};
    \path[fill=gray!3,rounded corners, draw=black!70, densely dotted]
      (a) rectangle (b);
  \end{pgfonlayer}
  \node [rectangle, above of=measurement, shift={(0.0,0.4)}, node distance=1.7em] (board) {\footnotesize Integrated circuit board};


\end{tikzpicture}
    }
    \caption{A diagram of the integrated distribution board. Thick lines in the diagram represent power connections, while thin lines show data connections.
    }
    \label{fig:integrated_board}
    \vspace{-1em}
\end{figure*}
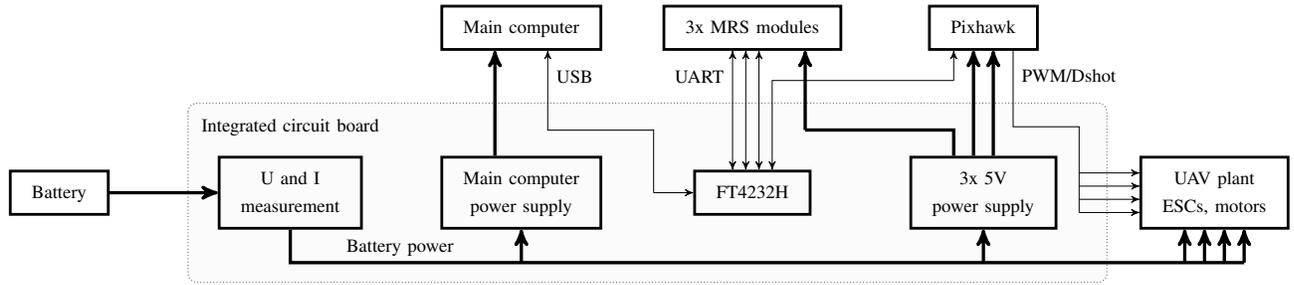

\subsection{Additional actuators}

The MRS UAV platforms are used in many applications, both indoor and outdoor. Such applications often require the use of specific (and sometimes unique) actuators. Among the actuators the MRS UAV platforms use are specialized grippers designed for grasping of metallic objects in the 2017 MBZIRC competition~\cite{spurny2019cooperative,loianno2018localization} and wall construction in the MBZIRC 2020~\cite{baca2020autonomous,stibinger2020mobile}. In MBZIRC 2020, a water cannon and fire blanket device for fire extinguishing was also used together with a system for catching a flying target~\cite{stasinchuk2020multiuav}.

The MRS platforms have also used actuators, such as manipulators attached to the UAVs~\cite{vrba2020autonomous}, gimbals for camera stabilization (NAKI project)~\cite{kratky2020autonomous, kratky2021documentation} and a capsule launcher for fire extinguishing (DOFEC project)~\cite{spurny2020autonomous}. In addition, even a net launcher, mounted on the Eagle.One\footnote{\url{https://eagle.one/en}} drone~\cite{vrba2020markerless}, was used to capture invading drones in aerial no-fly zones (for details see Sec.~\ref{sec:applications}).


\section{ELECTRICAL DESIGN AND CONFIGURATION}

The modular nature of the MRS UAV platform requires a lot of additional electronics,  such as low-level communication interfaces and power supplies for various sensors, lights and actuators. We designed a series of printed circuit distribution boards for the F450, T650 and X500 platforms, which are integrated into the platforms by replacing the top or the bottom structural board of the frames. The distribution board integrates two redundant power supplies for Pixhawk\footnote{The Pixhawk autopilot is an open-hardware and open-software architecture, which is advantageous for research in the field of aerial robotics~\cite{meier2015px4}.}, monitors the battery by measuring voltage and current, distributes power and throttle signals to the individual motors, and provides a standardized interface for other lower level boards, called MRS modules, which extend the capabilities of the platform. The center-point of these boards is a USB2quad serial converter, e.g., the FT4232H. It connects up to 4 separate UARTs through a single USB 2.0 cable to the main computer. One of the UARTs is reserved for communication between Pixhawk and the main computer, while the three remaining UARTs connect to the slots for MRS modules. One additional 5\,V power supply is powering the MRS modules. Functional diagram of the integrated distribution board is shown in  Fig. \ref{fig:integrated_board}.

\begin{figure}[tb]
    \centering
    \input{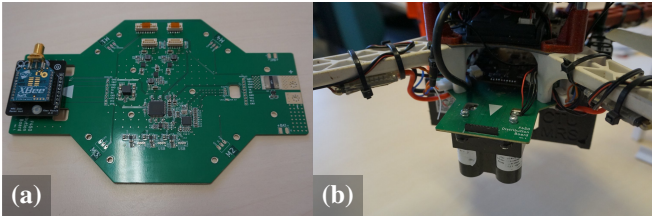}
    \vspace{-5mm}
    \caption{Distribution board for the F450 shown with an Xbee MRS module (a), and integrated into the F450 platform with a UVDAR controller MRS module (b).}
    \label{fig:distboard}
    \vspace{-1em}
\end{figure}

\begin{figure}[tb]
    \centering
    \includegraphics[width=1.0\columnwidth]{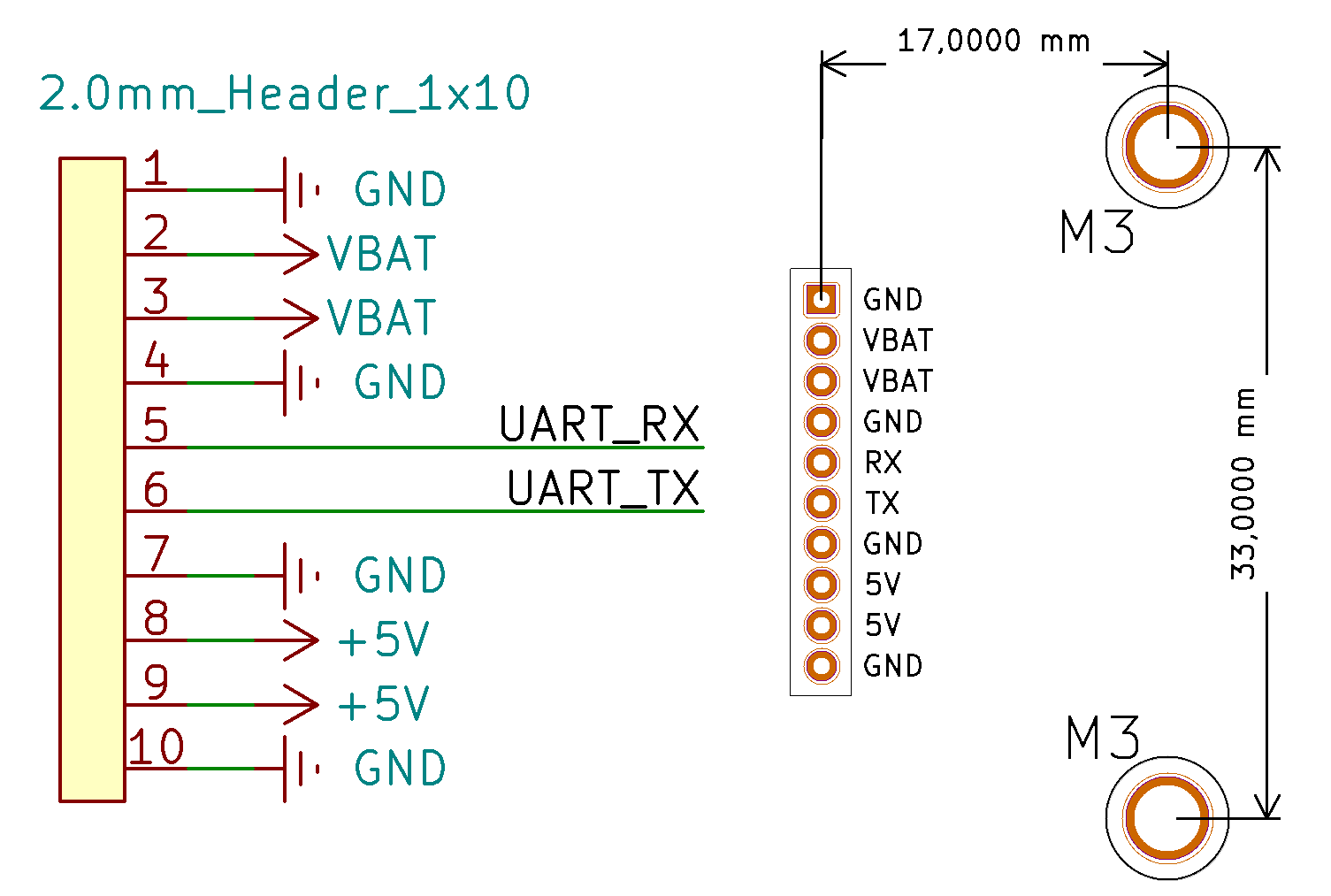}
    \vspace{-5mm}
    \caption{Standardized electrical and mechanical interface for MRS modules.}
    \label{fig:mrsmodule}
    \vspace{-1em}
\end{figure}

Up to three MRS modules can be installed to the main board to provide additional functionality. The MRS module is connected through a standard 2.0\,mm 10-pin header which provides a stabilized 5\,V power, direct connection to the battery for higher power applications and a UART communication interface to the main computer. The MRS modules are connected through the header and secured by two 5.0\,mm tall M3 mounting posts, in a standardized mounting pattern as shown in Fig.~\ref{fig:mrsmodule}.  Various MRS modules have been developed, for example a controller for the LEDs of the UVDAR system, interface for an Xbee radio (see Fig.~\ref{fig:distboard}) or a controller for LED strips.


\begin{figure*}[!t]
    \centering
    \resizebox{1.0\textwidth}{!}{
    \pgfdeclarelayer{foreground}
\pgfsetlayers{background,main,foreground}

\tikzset{radiation/.style={{decorate,decoration={expanding waves,angle=90,segment length=4pt}}}}

\begin{tikzpicture}[->,>=stealth', node distance=3.0cm,scale=1.0, every node/.style={scale=1.0}]


  \node[state, shift = {(0.0, 0.0)}] (navigation) {
      \begin{tabular}{c}
        \footnotesize Mission \&\\
        \footnotesize navigation
      \end{tabular}
    };


  \node[state, right of = navigation, shift = {(0.7, 0)}] (tracker) {
      \begin{tabular}{c}
        \footnotesize Reference \\
        \footnotesize tracker
      \end{tabular}
    };

  \node[state, right of = tracker, shift = {(0.1, 0)}] (controller) {
      \begin{tabular}{c}
        \footnotesize Reference \\
        \footnotesize controller
      \end{tabular}
    };

  \node[state, right of = controller, shift = {(0.8, -0)}] (attitude) {
      \begin{tabular}{c}
        \footnotesize Attitude rate\\
        \footnotesize controller
      \end{tabular}
    };

  \node[smallstate, below of = attitude, shift = {(-0.6, 2.1)}] (imu) {
      \footnotesize IMU
    };

  \node[state, right of = attitude, shift = {(0.7, -0)}] (actuators) {
      \begin{tabular}{c}
        \footnotesize UAV \\
        \footnotesize actuators
      \end{tabular}
    };

  \node[state, right of = actuators, shift = {(-0.8, -0)}] (sensors) {
      \begin{tabular}{c}
        \footnotesize Onboard \\
        \footnotesize sensors
      \end{tabular}
    };

  \node[state, below of = attitude, shift = {(0, 0.9)}] (estimator) {
      \begin{tabular}{c}
        \footnotesize State \\
        \footnotesize estimator
      \end{tabular}
    };

  \node[state, right of = estimator, shift = {(0.8, 0.0)}] (localization) {
      \begin{tabular}{c}
        \footnotesize Odometry \& \\
        \footnotesize localization
      \end{tabular}
    };



  \path[->] ($(navigation.east) + (0.0, 0)$) edge [] node[above, midway, shift = {(0.0, 0.05)}] {
      \begin{tabular}{c}
        \footnotesize desired reference\\
        \footnotesize $\mathbf{r}_d, \eta_d$\\
        \footnotesize \textit{on demand}
    \end{tabular}} ($(tracker.west) + (0.0, 0.00)$);


  \path[->] ($(tracker.east) + (0.0, 0)$) edge [] node[above, midway, shift = {(0.0, 0.05)}] {
      \begin{tabular}{c}
        \footnotesize full-state reference\\
        \footnotesize $\bm{\chi}_d$\\
        \footnotesize \SI{100}{\hertz}
    \end{tabular}} ($(controller.west) + (0.0, 0.00)$);

  \path[->] ($(tracker.south |- estimator.west) + (0.0, 0.0)$) edge [dotted] node[left, midway, shift = {(0.2, 0.00)}] {
      \begin{tabular}{r}
        \scriptsize initialization\\[-0.5em]
        \scriptsize only
    \end{tabular}} ($(tracker.south) + (0.0, 0.00)$);

  \path[->] ($(controller.east) + (0.0, 0)$) edge [] node[above, midway, shift = {(-0.2, 0.05)}] {
      \begin{tabular}{c}
        \footnotesize $\bm{\omega}_d$\\
        \footnotesize $T_d$ \\
        \footnotesize \SI{100}{\hertz}
    \end{tabular}} ($(attitude.west) + (0.0, 0.00)$);

  \draw[-] ($(controller.south)+(0.25,0)$) -- ($(controller.south |- estimator.west) + (0.25, 0.25)$) edge [->] node[above, near start, shift = {(-0.2, 0.05)}] {
      \begin{tabular}{c}
        \footnotesize $\mathbf{a}_d$
    \end{tabular}} ($(estimator.west) + (0, 0.25)$);

  \path[->] ($(attitude.east) + (0.0, 0)$) edge [] node[above, midway, shift = {(0.1, 0.05)}] {
      \begin{tabular}{c}
        \footnotesize $\bm{\tau}_d$ \\
        \footnotesize $\approx$\SI{1}{\kilo\hertz}
    \end{tabular}} ($(actuators.west) + (0.0, 0.00)$);

  \path[-] ($(estimator.west)+(0, 0)$) edge [] node[above, near start, shift = {(-1.1, 0.0)}] {
      \begin{tabular}{c}
        \footnotesize $\mathbf{x}$, $\mathbf{R}$, $\bm{\omega}$\\
        \footnotesize \SI{100}{\hertz}
    \end{tabular}} ($(navigation.south |- estimator.west)$) -- ($(navigation.south |- estimator.west)$) edge [->,] ($(navigation.south)+(0, 0)$);


  \path[->] ($(controller.south |- estimator.west)+(0, 0)$) edge [] ($(controller.south)$);

  \draw[-] ($(imu.east) + (0.0, 0.0)$) -- ($(estimator.north |- imu.east) + (0.3, 0)$) edge [->] node[right, midway, shift = {(-0.2, 0.3)}] {
      \begin{tabular}{c}
        \footnotesize $\mathbf{R}$, $\bm{\omega}$
    \end{tabular}} ($(estimator.north) + (0.3, 0.0)$);

  \draw[-] ($(sensors.south)+(0, 0)$) -- ($(sensors.south |- localization.east)$) edge [->] ($(localization.east)$);
  \draw[-] ($(sensors.south)+(0.1, 0)$) -- ($(sensors.south |- localization.east) + (0.1, -0.1)$) edge [->] node[midway, shift = {(0.0, -0.20)}] {
      \begin{tabular}{c}
    \end{tabular}} ($(localization.east) + (0.0, -0.1)$);
  \draw[-] ($(sensors.south)+(-0.1, 0)$) -- ($(sensors.south |- localization.east) + (-0.1, 0.1)$) edge [->]  ($(localization.east) + (0.0, 0.1)$);

  \draw[->] ($(localization.west)+(0, 0)$) -- ($(estimator.east)$);
  \draw[->] ($(localization.west)+(0, 0.1)$) -- ($(estimator.east) + (0, 0.1)$);
  \draw[->] ($(localization.west)+(0, -0.1)$) -- node[midway, shift = {(0.0, -0.2)}] {
      \begin{tabular}{c}
    \end{tabular}} ($(estimator.east) + (0, -0.1)$);




  \begin{pgfonlayer}{background}
    \path (attitude.west |- attitude.north)+(-0.45,0.8) node (a) {};
    \path (imu.south -| sensors.east)+(+0.25,-0.20) node (b) {};
    \path[fill=gray!3,rounded corners, draw=black!70, densely dotted]
      (a) rectangle (b);
  \end{pgfonlayer}
  \node [rectangle, above of=actuators, shift={(-0.6,0.55)}, node distance=1.7em] (autopilot) {\footnotesize UAV plant};

  \begin{pgfonlayer}{background}
    \path (attitude.west |- attitude.north)+(-0.25,0.47) node (a) {};
    \path (imu.south -| attitude.east)+(+0.25,-0.10) node (b) {};
    \path[fill=gray!3,rounded corners, draw=black!70, densely dotted]
      (a) rectangle (b);
  \end{pgfonlayer}
  \node [rectangle, above of=attitude, shift={(0,0.2)}, node distance=1.7em] (autopilot) {\footnotesize Embedded autopilot};


\end{tikzpicture}
    }
    \caption{Diagram of the system architecture: \emph{Mission \& navigation} software supplies the position and heading reference ($\mathbf{r}_d$, $\eta_d$) to a reference tracker.
\emph{Reference tracker} creates a smooth and feasible reference $\bm{\chi}$ for the feedback controller.
The feedback \emph{Reference controller} produces the desired thrust and angular velocities ($T_d$, $\bm{\omega}_d$) for the Pixhawk embedded flight controller.
The \emph{State estimator} fuses data from \emph{Onboard sensors} and \emph{Odometry \& localization} methods to create an estimate of the UAV translation and rotation ($\mathbf{x}$, $\mathbf{R}$).
    }
    \label{fig:system_architecture}
    \vspace{-1.2em}
\end{figure*}
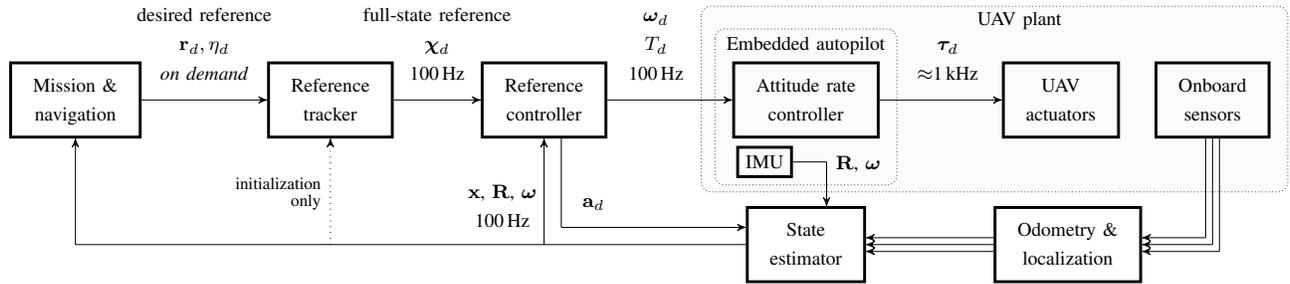




\section{MRS UAV system}


The presented hardware is complemented by the open-source MRS UAV System for control, estimation and deployment of multi-rotor aerial vehicles in realistic simulations and real-world scenarios~\cite{baca2021mrs}.
The system enables complex missions in GNSS and GNSS-denied environments, including outdoor-indoor transitions and the execution of redundant estimators for backing up unreliable localization sources.
Two feedback control designs are used: one for precise and aggressive maneuvers, and the other for stable and smooth flight with a noisy state estimate.
The control reference generation is provided by the unique real-time virtual Model Predictive Control (MPC) tracker~\cite{baca2018model}.
Although the system comes with well-tested and real-world proven control pipeline, it is also modular.
The system allows the users to safely develop and test novel feedback controllers, reference generators and estimators with the ability to switch back to the baseline implementation.
The MRS UAV control and estimation pipelines are constructed without using the Euler/Tait-Bryan angle representation of orientation in 3D, that only causes confusion due to the ambiguities and singularities.
Instead, we rely on rotation matrices and a novel heading-based convention to represent the one free rotational degree-of-freedom in 3D of a standard multi-rotor aircraft.
We provide an actively maintained and well-documented open-source implementation, including realistic simulation of UAVs, sensors, and localization systems.
The MRS UAV system, with a pipeline used for control and estimation depicted in Fig.\,\ref{fig:system_architecture}, has been used in various real-world system deployment that subsequently shaped the system into the form presented here.




\section{Applications for the Aerial Platforms}
\label{sec:applications}

Despite the differences in the presented hardware platforms, the hardware and software stacks of the MRS UAV system are coupled such that the software copes easily with hardware irregularities as well as modularities.
This is an important feature from the viewpoint of wide general research.
In the following subsections, the aerial platforms are introduced in context of realistic Software-In-The-Loop (SITL) simulations, research on aerial autonomy in indoor and outdoor environments, robotic competitions, and prototyping for industrial applications.
The primary attributes of example platforms are summarized in Table~\ref{tab:platforms}.  


\begin{table}[!t]
  \centering
  \sisetup{per-mode=symbol}
  \caption{Aerial platforms utilizing the MRS software stack in research, academic and industry projects. \textit{Dimensions} is represented by length of the main diagonal without propellers. \textit{Parts} denotes \textbf{P}ublicly available and \textbf{C}ustom-made parts required for construction. \textit{Purpose} denotes \textbf{R}esearch and \textbf{I}ndustrial platforms.%
       \vspace{-1mm}
       }
  \resizebox{\columnwidth}{!}{%
  \begin{tabular}{l l l l l l l}
    \toprule
    Platform              & X500  & F450 & T650 & NAKI & Eagle.One & DOFEC\\\midrule
    Flight time (min)     & 25    & 15   & 20   & 7    & 10        & 10\\
    Weight (kg)           & 2.5   & 1.7  & 3.5  & 5.5  & 10        & 7\\
    Dimension (mm)        & 500   & 450  & 650  & 570  & 1250        & 657\\
    Propeller size (in)   & 13 & 9.4  & 15   & 12   & 18        & 15\\
    Battery capacity (Wh) & 199.8  & 99.9 & 177.6    & 355.2  & 355.2        & 355.2 \\
    Rotors count          & 4     & 4    & 4    & 8    & 8         & 8\\
    Parts                 & P     & P    & P    & C    & C         & C\\
    Purpose               & R/I   & R    & R    & R    & I         & I\\
    \bottomrule
  \end{tabular}
  }
  \label{tab:platforms}
  \vspace{-1.5em}
\end{table}



\subsection{Gazebo realistic simulations}


We have developed a simulation environment, which was made publicly available\footnote{\url{https://github.com/ctu-mrs/simulation}}, to facilitate multi-UAV experiments.
It makes use of the open-source Gazebo simulator and is set up for multiple different variants of our hardware \ac{UAV} platforms (DJI F450, DJI F550, Tarot 650 sport, etc.).
It can be also easily extended to a new hardware setup once required.
All \ac{UAV} hardware elements, including the Pixhawk flight controller, various sensors, and actuators are simulated with high fidelity, to minimize difference between simulated flight and real-world flight. 
This ensures a smooth transition between simulation and reality, which significantly accelerates the deployment of new robotic methods and algorithms.
Therefore, hardware experiments can be realized in a shorter time and with fewer safety risks than relying on direct hardware verification. 



\subsection{Indoor real robot experiments}

In this section, we demonstrate in multiple scenarios that with a rapid sensory and actuation modification, the \ac{UAV} is able to localize itself in an indoor environment and achieve the desired mission objectives.



\subsubsection{DARPA SubT --- S\&R competition and exploration of subterranean environments}
 
Aiming to speed up research of autonomous Search \& Rescue operations in underground environments, DARPA has organized the Subterranean Challenge\footnote{\url{http://mrs.felk.cvut.cz/projects/darpa}}. 
Participating teams had to develop a robotic solution capable of navigating in an unknown environment, detecting survivors and their belongings, and reporting their exact positions.      
The deployment of the UAV platform in harsh, constrained, and unknown environments imposes mutually competing requirements on minimization of UAV dimensions while preserving long flight time and extensive payload.
An omnidirectional sensory setup composed of wide-angle 3D LiDAR and two depth cameras that cover the blind spots of the LiDAR above and below the UAV enable safe exploration of completely unknown environment in horizontal as well as vertical directions.

\begin{figure}[tb]
    \centering
    \input{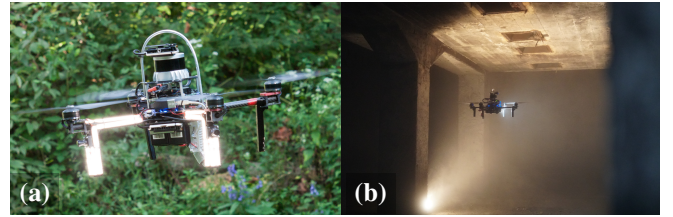}
    \vspace{-5mm}
    \caption{Platform (a) designed for the exploration of subterranean environments (b). Video: \url{https://youtu.be/WG3CthG6XuU}.}
    \label{fig:darpa}
    \vspace{-1.5em}
\end{figure}

The additional sensors (RGB cameras coupled with onboard lights, thermal cameras, and sensors for identifying gas leaks) are employed for perception of survivors and their belongings. 
All these components are assembled on a frame in a compact way, minimizing the weight of the payload and overall platform dimensions, as depicted in Fig.~\ref{fig:darpa}.
The results achieved during deployment of the platform in DARPA SubT Challenge are described in~\cite{petrlik2020robust, kratky2021exploration, petracek2021caves, Pritzl2022ICUAS}.
The platform was designed to allow agile flights with velocities reaching up to \SI{8}{\meter\per\second}, while keeping sufficiently low dimensions necessary for traversal of narrow passages typical for man-made subterranean environments such as doors and windows, or for cavity entrances in natural structures.
Vastness of such environments is also tackled by flight-time of over \SI{20}{\minute}. In addition,
the UAV design allows for easy imitation of the platform with simulation model achieving accurate real-world performance.
The simulation model based on real UAV performance was used in the virtual track of the DARPA SubT Finals by majority of participating teams, including the winners.
Apart from the participation in DARPA SubT, where our robotic team consisting of 5 UAVs and 2 UGVs achieved second place in the virtual track, the platform was also subjected to extensive tests in environments of varying characteristics including natural caves, underground fortresses, mines, and cluttered outdoor environments.   

\subsubsection{Dronument: documentation of historical monuments}

\begin{figure}[tb]
    \centering
    \input{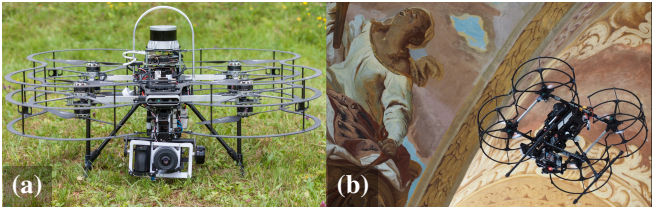}
    \vspace{-5mm}
    \caption{Platform (a) tailored for documentation of structures and valuables within interiors of historical monuments (b). Video: \url{https://youtu.be/-_1Fjr58a28}.}
    \label{fig:dronument}
\end{figure}

The MRS hardware stack includes platforms designed for autonomous and cooperative high-resolution photography in interiors of buildings.
The dimensions of these platforms are minimized, whereas robustness and reliability are maximized under constraints on critical safety.
In addition to research platforms, the safety-critical branch of the hardware stack provides redundant sensory equipment to maximize the on-board perception and hardware solutions to prevent collisions of the propellers with the environment.
The extra sensors include a set of wide-field-of-view ultrasonic sensors, mounted in an omnidirectional manner and utilized in low-level supervision of the primary LiDAR sensor used for localization and mapping.
The nature of laser-based and ultrasonic-based measurements makes their coupling complementary for proximal obstacle detection.
For operation purposes, mission diagnostics and failure detection systems are combined with information about proximity to obstacles and are encoded into a visual health status visible to a human supervisor through a powerful on-board LED, hence warning about a possible danger.
Utilization of the safety-critical platforms in tasks documenting interiors of historical monuments by a team of \acp{UAV} is part of the Dronument\footnote{\url{http://mrs.felk.cvut.cz/dronument}} project, showcased in Fig.~\ref{fig:dronument} and summarized in~\cite{saska2017documentation,petracek2020dronument,kratky2020autonomous,kratky2021documentation,smrcka2020admittance, Bednar2022ICUAS}.

\subsubsection{Industrial inspection}

\begin{figure}[tb]
    \centering
    \input{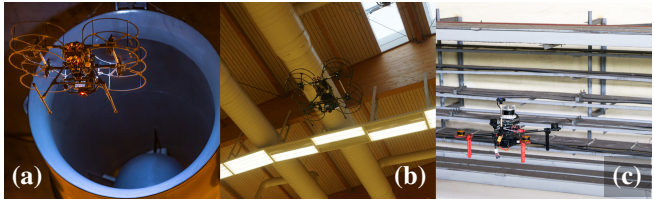}
    \vspace{-5mm}
    \caption{Snapshots of deployment of industrial platforms (a,~b) and a general research platform (c) for inspection of ventilation systems, structural degradation, and electrical infrastructures in tunnels, halls, and storage houses. Video: \url{https://youtu.be/60nKXamV2ds}.}
    \label{fig:industry_indoor_inspection}
    \vspace{-2em}
\end{figure}

The platforms designed for interior documentation are likewise applicable in inspection tasks within industrial structures --- factory halls, storage houses, and others.
Characteristics of these environments include high structurality with repetitive featureless patterns (for both visual and laser sensing) and presence of many small-scale hard-to-detect obstacles (ropes, cables, and others).
These attributes yield harsh requirements on the level of on-board perception.
High-level perception required for industrial scenarios can be achieved with the platforms developed for interior documentation, which feature high payload capacities exploitable for perceptual enhancement.
Example scenarios are showcased in Fig.~\ref{fig:industry_indoor_inspection}.

\subsection{Outdoor real robot experiments}

In outdoor experiments, the MRS platforms may fully rely on GNSS data, but can also smoothly transit to GNSS-denied regions (e.g., forests). In this subsection we demonstrate the adaptability of the MRS platforms to be used in a diverse range of scenarios and applications, including swarming in desert and forest environments, power-line monitoring, and human-robot interaction.



\subsubsection{Swarming in desert, hills and forest}

\begin{figure}[tb]
    \centering
    \input{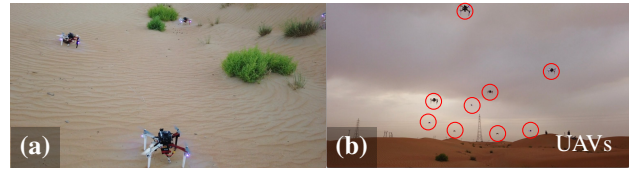}
    \caption{Swarms of UAVs in the desert (a) using MRS platform and 3D formation of 10 UAVs (b). Video: \url{https://youtu.be/o8bphtbPCaA}.}
    \label{fig:swarms}
\end{figure}

\begin{figure}[tb]
    \centering
    \input{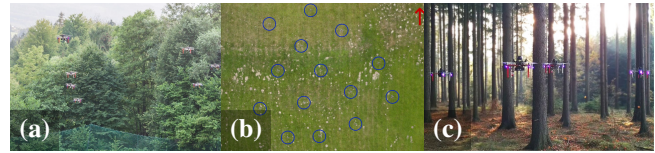}
    \vspace{-5mm}
    \caption{Swarms of UAVs in the grass hill (a, b) and forest environments (c). Video: \url{https://youtu.be/HH78AheC-DM}.}
    \label{fig:swarms2}
    \vspace{-1em}
\end{figure}

Swarm control is aimed at coordinated motion of a large group of individuals moving together towards the same target direction, as depicted in Fig.~\ref{fig:swarms}. This collective behavior can be observed in different species in nature. In robotics, different methods have been proposed to accomplish flocking behavior in multi-robot systems \cite{RODRIGUEZ2021,WU2022}. 
To achieve swarming of aerial robots in real-world conditions, MRS platforms rely on the UVDAR system for keeping cohesion and avoiding collisions between agents. The first deployment of self-stabilised UAV swarms without any communication was presented in \cite{petracek2020bioinspired}. A compact group of MRS UAVs was deployed in a desert environment \cite{ICUAS_Thulio, Krizek2022ICUAS} for search and rescue applications with CNN camera detector for human-victim detection in a control loop. These experiments also verified MRS platforms under demanding light and temperature conditions. A similar experimental verification was performed at a grass hill, as shown in Fig.~\ref{fig:swarms2}, where the swarm system showed the ability to follow uneven terrain. In works \cite{afzal2021Icra, dmytruk2020safe, Akash2022ICUAS}, a UAV swarm was deployed in a demanding forest environment. The compact group of robots was able to safely navigate through dense obstacle area using 2D LiDAR. In \cite{novak2021fast}, a bio-inspired evasion approach in self-organized swarm of UAVs was introduced.  Using the UVDAR system, MRS UAVs were able to avoid dynamic objects (predators) that were actively approaching the group.

\subsubsection{MBZIRC 2017 and 2020}

The MBZIRC 2017\footnote{\url{http://mrs.felk.cvut.cz/mbzirc}} and 2020\footnote{\url{http://mrs.felk.cvut.cz/mbzirc2020}} competitions were composed of several challenges motivated by the intent to push technological and application boundaries in robotics beyond the current state-of-the-art. 
These technological challenges include fast autonomous navigation in semi-unstructured, complex, and dynamic environments with minimal prior knowledge, robust perception and tracking of dynamic objects in 3D, sensing and avoiding obstacles, GNSS denied navigation in indoor-outdoor environments, physical interactions, complex mobile manipulations, and air-surface collaboration.
In MBZIRC 2017, the system relying on MRS platforms won Challenge 3, where a team of three UAVs had to find a set of static and moving colored ferromagnetic objects and deliver them to the target location\footnote{\label{footmbzirc}\url{https://youtu.be/ogmQSjkqqp0}}~\cite{spurny2019cooperative,loianno2018localization}.
Furthermore, the system achieved 2nd place in Challenge 1, where a single UAV had to autonomously localize a moving vehicle in the arena and land on  the vehicle\footref{footmbzirc}~\cite{baca2019autonomous,stepan2019vision}. 
In MBZIRC 2020, the system using MRS platforms achieved 1st place in Challenge~2 --- autonomous wall structure building by the team of three UAVs\footnote{\url{https://youtu.be/1-aRtSarYz4}} and one UGV~\cite{baca2020autonomous,stibinger2020mobile}, and 2nd place in Challenge~1, where UAVs should autonomously track and interact with a flying target\footnote{\url{https://youtu.be/2-cLSjRCKDg}}~\cite{vrba2020autonomous,stasinchuk2020multiuav,stasinchuk2020fr}.
The MRS UAV system had also been applied in the fire-fighting\footnote{\url{https://youtu.be/O8QBiAyP2c0}} Challenge~3 of MBZIRC 2020~\cite{jindal2020fr,walter2020fr,walter2020icuas,spurny2020autonomous}. 

\begin{figure}[tb]
    \centering
    \includegraphics[width=0.20\columnwidth]{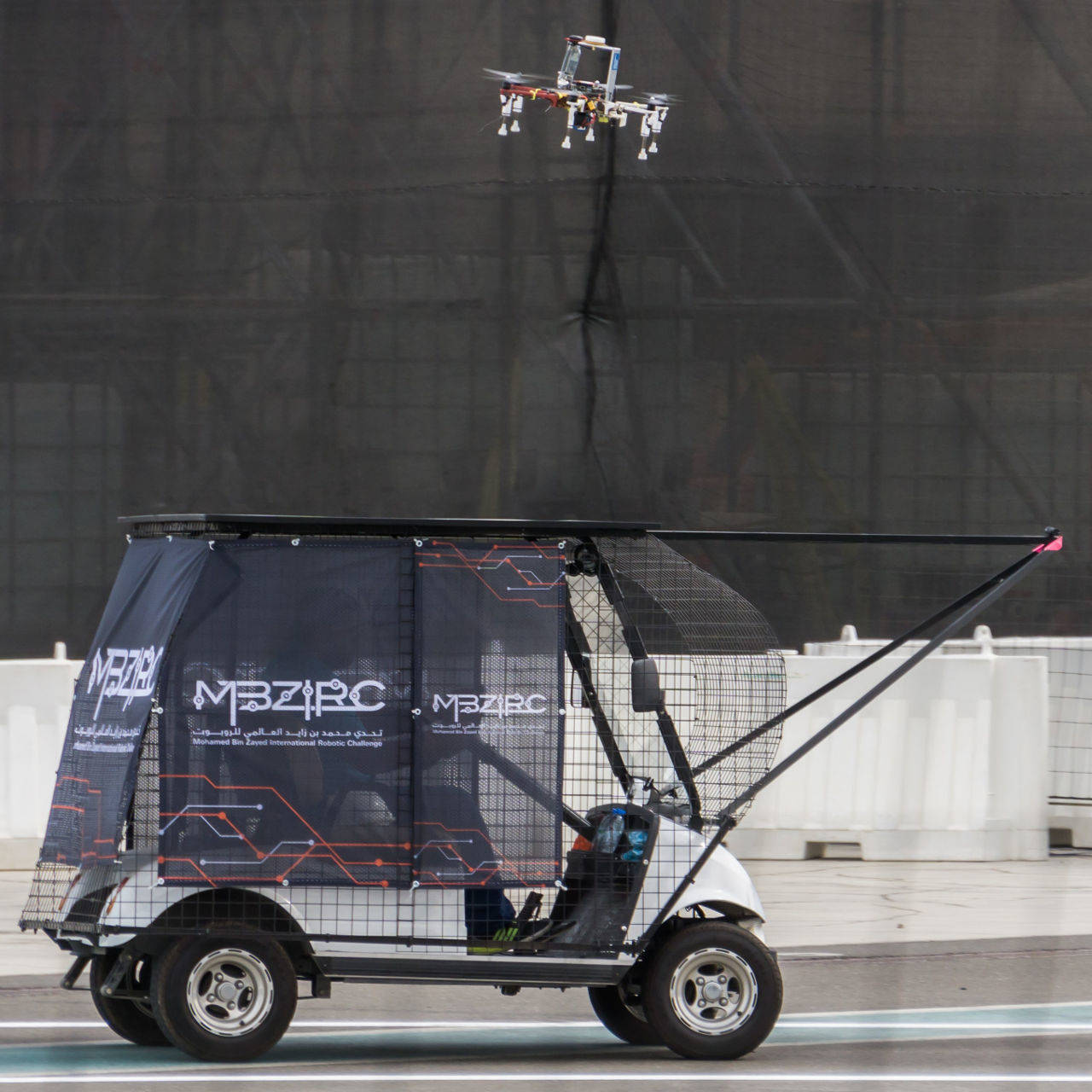}
    \includegraphics[width=0.20\columnwidth]{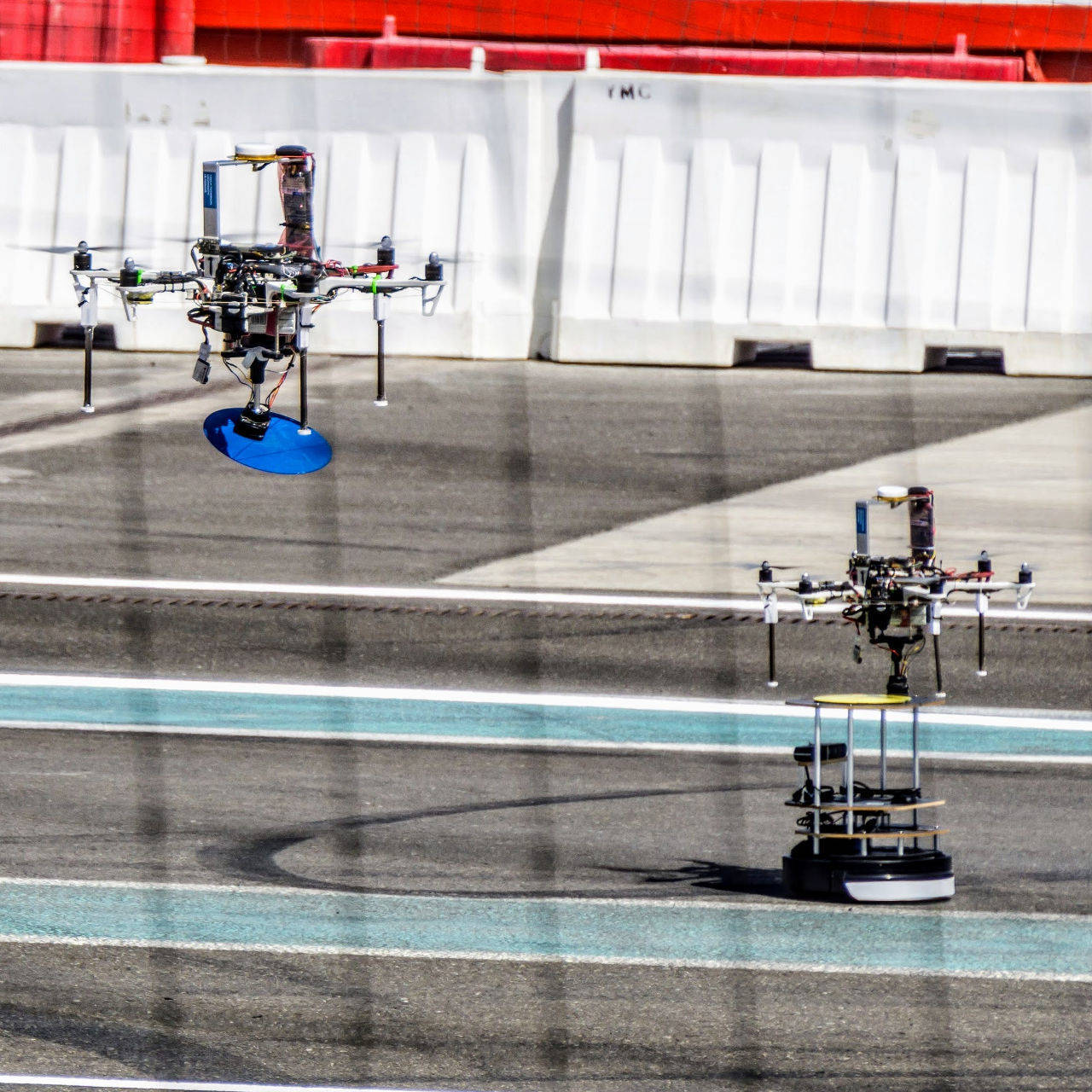}
    \includegraphics[width=0.17\columnwidth]{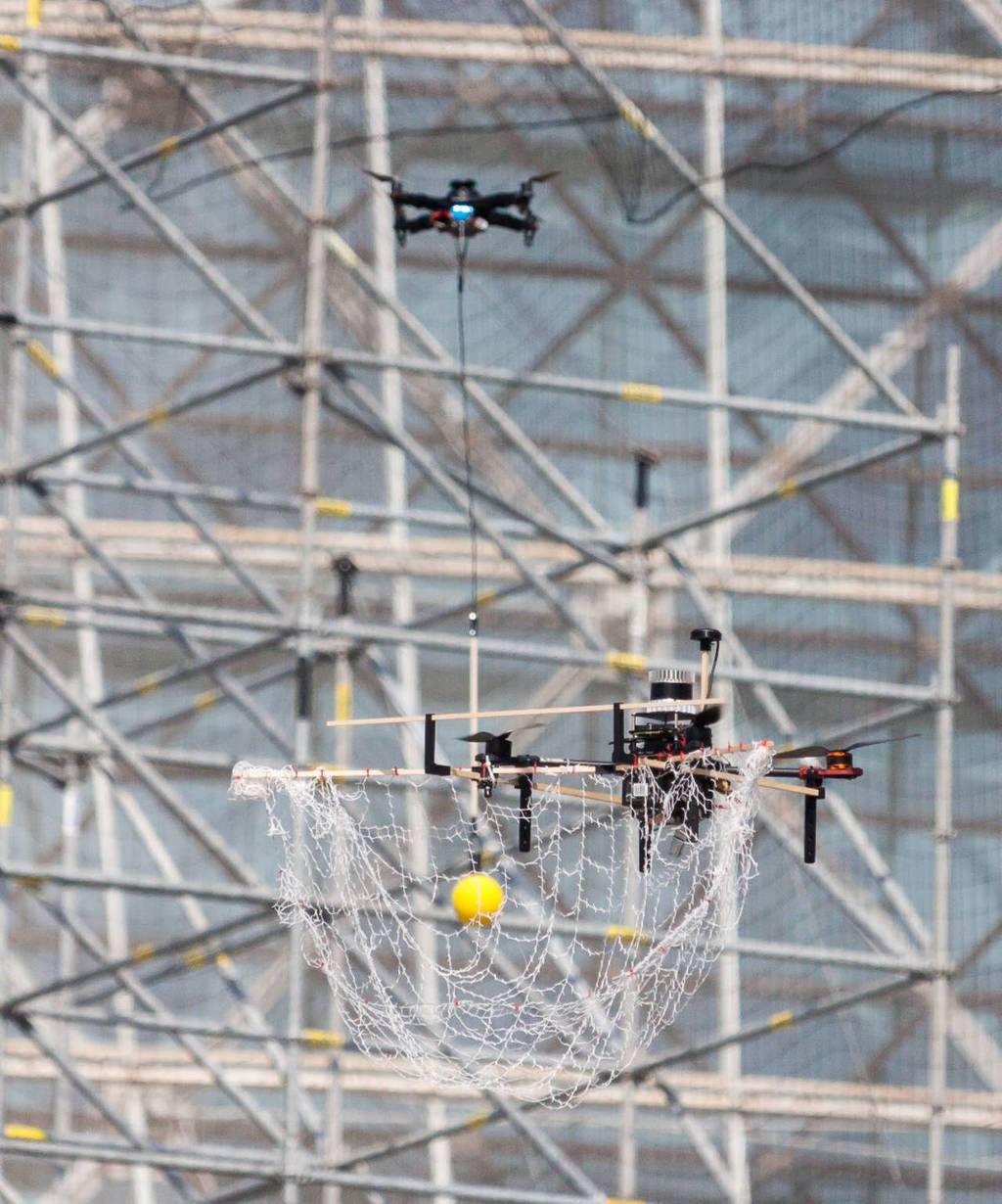}
    \includegraphics[width=0.17\columnwidth]{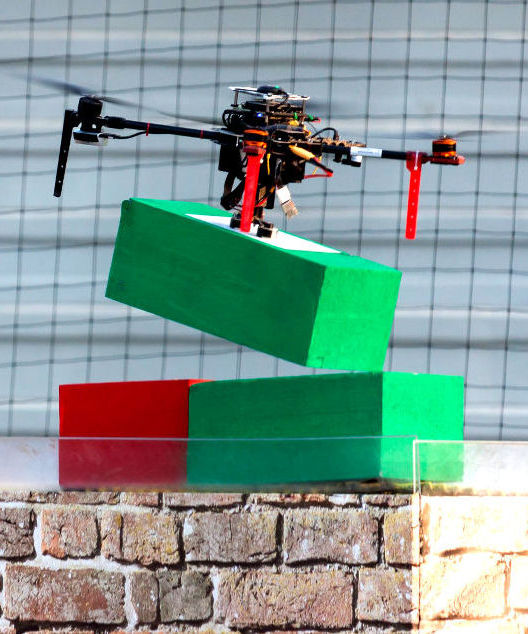}
    \includegraphics[width=0.17\columnwidth]{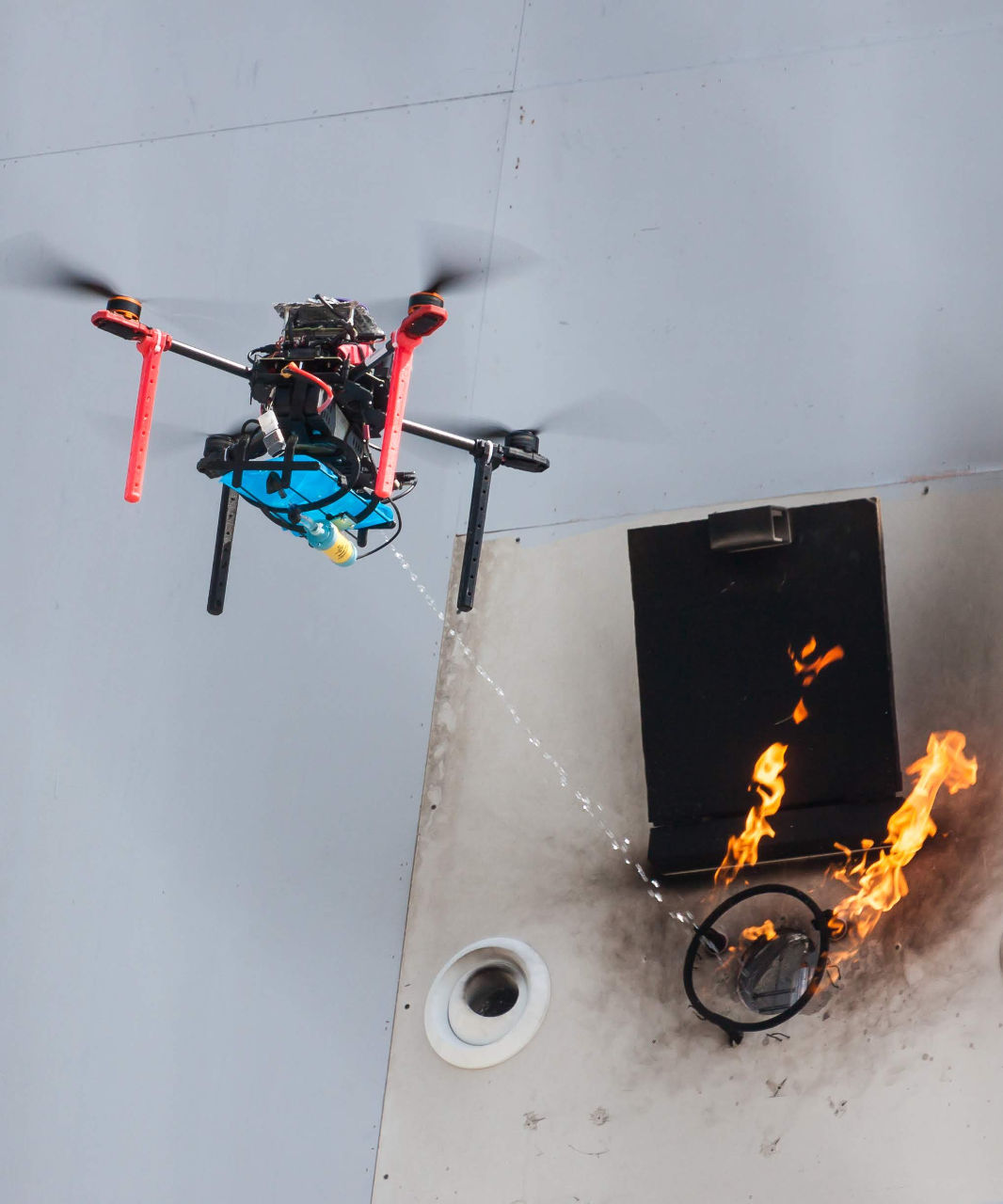}
    \caption{MBZIRC 2017 (autonomous landing on a moving vehicle and cooperative collection of objects) \& MBZIRC 2020 (autonomous capture of agile objects, cooperative wall building and autonomous fire-fighting). Video: \url{https://youtu.be/DEUZ77Vk2zE}.}
    \label{fig:dofec}
    \vspace{-1.5em}
\end{figure}



\subsubsection{Aerial-Core}
\label{sec:aerialCore}

\begin{figure}[tb]
    \centering
    \input{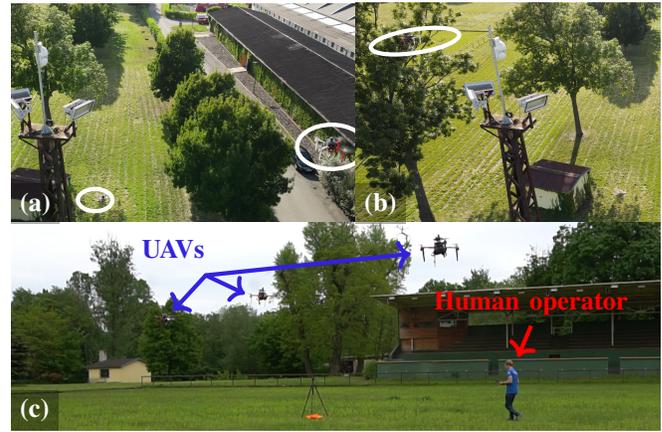}
    \caption{Snapshots of the \ac{MRS} drones in the \textit{inspection} and \textit{monitoring} operations withing the AERIAL-CORE European project. Figures (a) and (b) show the \textit{inspection} scenario. Solid circles show the \acp{UAV} approaching the tower. Fig.~(c) shows the \textit{monitoring} scenario with the \acp{UAV} providing assistance to the human operator.}
    \label{fig:aerialCoreSnapshots}
    \vspace{-1em}
\end{figure}

The results presented in this subsection were achieved within the AERIAL-CORE\footnote{\url{https://aerial-core.eu}} European project. This project aims at developing cognitive aerial platforms inspired by the application of autonomous power line inspection. Two tasks of interest are considered: (i) \textit{inspection}, where a fleet of~\acp{UAV} carries out a detailed investigation of power equipment, assisting human operator in acquiring views of the power tower that are not easily accessible, as depicted in Fig.~\ref{fig:aerialCoreSnapshots}; (ii) \textit{monitoring}, where a formation of~\acp{UAV} provides to the supervising team a view of the humans working on the power tower to monitor their status and ensure their safety, as shown Fig.~\ref{fig:aerialCoreSnapshots}. 
In both tasks, visual sensors are essential to perform the assigned tasks. In the~\ac{UAV} configuration, cameras are mounted in \textit{eye-in-hand} configuration, i.e., rigidly attached to the body frame of the aircraft. Cameras are also used to mutually localize the~\acp{UAV} in the surrounding environment.
Further details about tasks and the designed algorithms and software architecture can be found in~\cite{Silano2021RAL, Kratky2021RAL, Silano2021ICUAS, Demkiv2021AIRPHARO, Nekovar2021RAL, Licea2021EUSIPCO, Calvo2022ICUAS, Silano2022ICUAS}. Illustrative videos of the experiments using the MRS UAV platforms are available\footnote{\url{https://mrs.felk.cvut.cz/projects/aerial-core}}.

\subsection{Industrial collaboration}

MRS hardware and software platforms kick-started several industrial projects, serving as a basis for preliminary proof-of-concept prototypes and tests. These collaborations were followed by further development of specialized platforms.

\subsubsection{Airspace protection}
\begin{figure}[tb]
    \centering
    \input{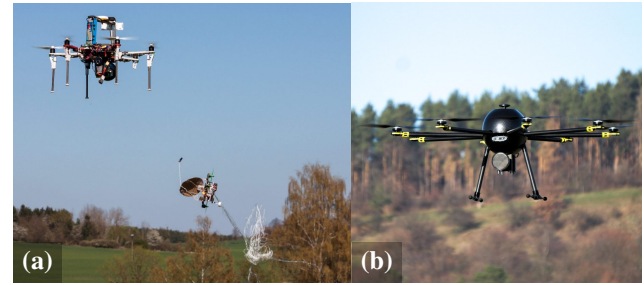}
    \caption{Development of the \textit{Eagle.One} autonomous aerial interception system based on the MRS hardware platform: (a) First prototype using the MRS UAV platform during a successful interception.  (b) Second prototype designed by~the~MRS team specifically for \textit{Eagle.One} project. Video: \url{https://youtu.be/hEDGE7ofX1c}.}
    \label{fig:eagle}
    \vspace{-1em}
\end{figure}

Example of a platform developed as a part of an industrial collaboration is the \textit{Eagle.One} system designed for autonomous aerial interception of intruder \acp{UAV}~\cite{vrba2019onboard, vrba2020autonomous}.
The first hardware iterations of the interceptor prototype were built directly using the basic MRS platforms until a dedicated platform with MRS system was designed, as shown in Fig.~\ref{fig:eagle}.
Initially during the development, the target was detected using a stereo camera and a specialized onboard net-launcher was used to capture it~\cite{vrba2019onboard}.
The design later converged to a powerful octo-rotor platform equipped with a 3D LiDAR sensor for detection of the target and a deployable net that is suspended below the interceptor platform and serves to catch the target.
The current design has a carry capacity of \SI{12}{\kilo\gram} to lift the most commonly available commercial platforms with a sufficient margin for dynamic maneuvering.

\subsubsection{Fire extinguishing}

Autonomous extinguishing of fires located in multi-floor buildings is tackled by a custom platform \textit{DOFEC}\footnote{\href{http://mrs.felk.cvut.cz/projects/dofec}{\url{mrs.felk.cvut.cz/projects/dofec}}} with MRS system.
This heavy-load octo-rotor platform carries perceptual sensors and a launcher loaded with a capsule filled with a fire-extinguishing substance.
After launching the capsule into the thermal source, detected by thermal camera and localized by RGB-D cameras on-board, the control systems of the platform autonomously copes with the dynamic recoil. 
The platform is capable of autonomous GNSS-enabled flight in proximity of tall buildings, as showcased in Fig.~\ref{fig:dofec}.

\begin{figure}[tb]
    \centering
    \includegraphics[width=1.0\columnwidth]{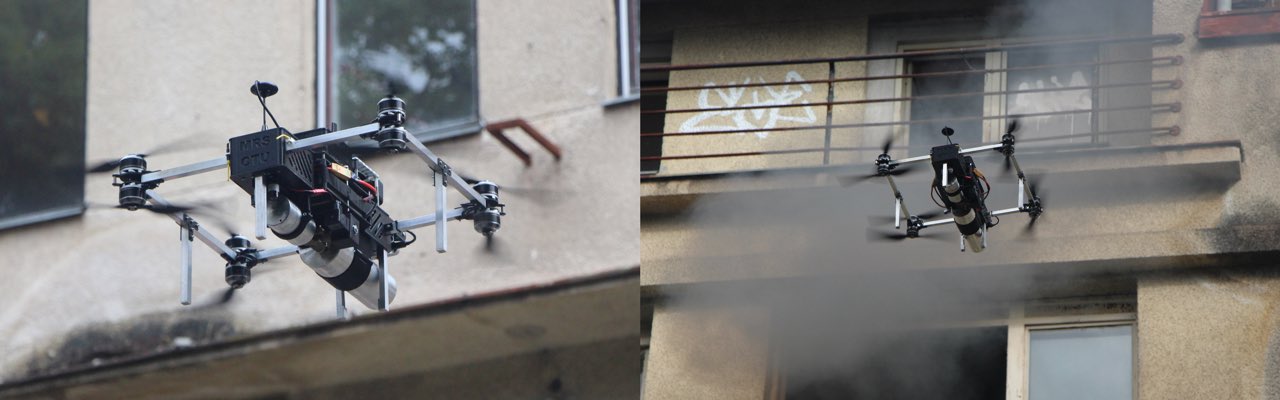}
    \vspace{-1.0em}
    \caption{Extinguishment of fires in above-ground floors by launching fire-extinguishing capsules into the thermal sources.
             Multimedia is available at~\url{https://youtu.be/QHpifXJzH5g}.}
    \label{fig:dofec}   
    \vspace{-1.0em}
\end{figure}


\subsubsection{Localization of radiation sources}

\begin{figure}[tb]
    \centering
    \input{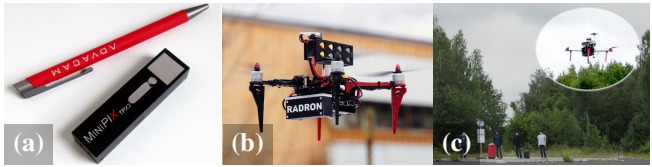}
    \vspace{-1.0em}
    \caption{Miniature Compton Camera MiniPIX Timepix3 (a) used as an onboard radiation sensor for ionizing radiation localization by \acp{UAV} (b,c). Video: \url{https://youtu.be/oH4jMMHfGVA}.}
    \label{fig:radron}
    \vspace{-1.0em}
\end{figure}

Ionizing radiation poses an invisible threat to humans and also to the environment.
Through our collaboration in project RaDron\footnote{\url{http://mrs.felk.cvut.cz/projects/tacr-radron-project}}, we have developed a compact aerial system for systematic radiation surveillance, mapping, and fast proactive localization of radiation sources \cite{stibinger2020localization, baca2021icuas}.
The platform is equipped with a cutting-edge single-detector Compton camera MiniPIX Timepix3 \cite{turecek2020single, baca2019timepix}, which is rigidly mounted to the MRS \ac{UAV} as a forward-facing camera, as depicted in Fig.\ref{fig:radron}.
The detector enables real-time estimation of the direction towards the source as well as the radiation intensity.
Similarly to the DARPA SubT challenge, the platform is equipped with a 3D LiDAR which enables operation in GNSS-denied environments, such as mine shafts, forests and underground parking.

\section{CONCLUSIONS}

This paper presented hardware platform designed for conducting research of UAVs in real-world conditions. 
We showed, that the proposed design is modular and allows to achieve a proper experimental setup for single, as well as, multi-robot scenarios using various actuators, sensors, and even UAV frames.
The experience and knowledge gained during thousands of hours of deployment of individual UAVs, as well as UAV teams in demanding outdoor and indoor conditions are described in this paper. 
Furthermore, we detailed recommendations and technical details required to start with prototyping and designing UAV platforms for research and validation of research hypotheses. 
Although the presented list of environments and tasks in which the proposed system was employed is relatively long (and not complete), the paper is intended to facilitate UAV prototyping in fully autonomous missions going beyond this list using a large number of components and their possible combinations. 

\begin{acronym}
  \acro{CNN}[CNN]{Convolutional Neural Network}
  \acro{IR}[IR]{infrared}
  \acro{GNSS}[GNSS]{Global Navigation Satellite System}
  \acro{MOCAP}[mo-cap]{Motion capture}
  \acro{MPC}[MPC]{Model Predictive Control}
  \acro{MRS}[MRS]{Multi-robot Systems group}
  \acro{ML}[ML]{Machine Learning}
  \acro{MAV}[MAV]{Micro-scale Unmanned Aerial Vehicle}
  \acro{UAV}[UAV]{Unmanned Aerial Vehicle}
  \acro{UV}[UV]{ultraviolet}
  \acro{UVDAR}[\emph{UVDAR}]{UltraViolet Direction And Ranging}
  \acro{UT}[UT]{Unscented Transform}
  \acro{RTK}[RTK]{Real-Time Kinematic}
  \acro{ROS}[ROS]{Robot Operating System}
  \acro{wrt}[w.r.t.]{with respect to}
\end{acronym}


\balance
\bibliographystyle{IEEEtran}
\bibliography{bib_short} 

\end{document}